\documentclass[11pt,letterpaper]{article}
\usepackage{acl2010}
\usepackage{times}
\usepackage{latexsym}

\usepackage{helvet}
\usepackage{courier}
\usepackage{times}
\usepackage{url}
\usepackage{latexsym}
\usepackage{multirow}
\usepackage{algorithm}
\usepackage{algorithmic}
\usepackage{graphics}
\usepackage{graphicx}
\usepackage{epsfig}
\usepackage{amssymb}
\usepackage{amsmath}
\usepackage{tablefootnote}
\usepackage{listings}

\DeclareMathOperator*{\argmax}{\arg\!\max}

\makeatletter
\newcommand*{\rom}[1]{\expandafter\@slowromancap\romannumeral #1@}
\makeatother

\def\ignore#1{}

\def\ignore#1{}

\graphicspath{{figures/}}

\def\ignore#1{}

\author{Peng Li \\
  Computer Science and Engineering \\
  University of Texas at Arlington \\
  {\tt jerryli1981@gmail.com} \\\And
  Heng Huang \\
  Computer Science and Engineering \\
  University of Texas at Arlington \\
  {\tt heng@uta.edu} \\}

\begin{document}

\title{Clinical Information Extraction via Convolutional Neural Network}
\maketitle

\begin{abstract}
We report an implementation of a clinical information extraction tool that leverages deep neural network to annotate event spans and their attributes from raw clinical notes and pathology reports. Our approach uses context words and their part-of-speech tags and shape information as features. Then we hire temporal (1D) convolutional neural network to learn hidden feature representations. Finally, we use Multilayer Perceptron (MLP) to predict event spans. The empirical evaluation demonstrates that our approach significantly outperforms baselines.
\end{abstract}

\section{Introduction}
In the past few years, there has been much interest in applying neural network based deep learning techniques to solve all kinds of natural language processing (NLP) tasks. From low level tasks such as language modeling, POS tagging, named entity recognition, and semantic role labeling~\cite{Collobert2011,Mikolov2013}, to high level tasks such as machine translation, information retrieval, semantic analysis~\cite{kalchbrenner2013,Socher2011dynamic,Tai2015} and sentence relation modeling tasks such as paraphrase identification and question answering~\cite{Socher2011,Mohit2014,Yin2015Multi}. Deep representation learning has demonstrated its importance for these tasks.  All the tasks get performance improvement via learning either word level representations or sentence level representations. 

In this work, we brought deep representation learning technologies to the clinical domain. Specifically, we focus on clinical information extraction, using clinical notes and pathology reports from the Mayo Clinic. Our system will identify event expressions consisting of the following components:
\begin{itemize}
\item The spans (character offsets) of the expression in the raw text
\item Contextual Modality: ACTUAL, HYPOTHETICAL, HEDGED or GENERIC
\item Degree: MOST, LITTLE or N/A
\item Polarity: POS or NEG
\item Type: ASPECTUAL, EVIDENTIAL or N/A
\end{itemize}

The input of our system consists of raw clinical notes or pathology reports like below:

\begin{description}
\item[] \emph{~~~~~~April 23, 2014: The patient did not have any postoperative bleeding so we will resume chemotherapy with a larger bolus on Friday even if there is slight nausea.}
\end{description}

And output annotations over the text that capture the key information such as event mentions and attributes. Table~\ref{tab:extract} illustrates the output of clinical information extraction in details. 

\begin{table*}[t!]
\begin{center}
\begin{tabular}{c|p{2.5cm}|p{9cm}}
\hline
Clinical Note & Event Mention & Event Attribute \\
\hline
\multirow{5}{3cm}{April 23, 2014: The patient did not have any postoperative bleeding so we will resume chemotherapy with a larger bolus on Friday even if there is slight nausea.} & \multirow{2}{*}{bleeding} &  type=N/A \hspace{1cm} polarity=NEG \\
                               		      				         &                                         &  degree=N/A \hspace{1cm} modality=ACTUAL \\
\cline{2-3}
              				     					 & \multirow{2}{*}{resume} &  type=ASPECTUAL \hspace{1cm} polarity=POS \\
                                                  					&                                       &  degree: N/A \hspace{1cm} modality=ACTUAL \\
\cline{2-3}
              				     					 & \multirow{2}{*}{chemotherapy} &  type=ASPECTUAL \hspace{1cm} polarity=POS \\
                                                  					 &                                       &  degree=N/A \hspace{1cm} modality=ACTUAL \\
\cline{2-3}
              				     					 & \multirow{2}{*}{bolus} &  type=ASPECTUAL \hspace{1cm} polarity=POS\\
                                                  					 &                                       &  degree=N/A \hspace{1cm} modality=ACTUAL \\
\cline{2-3}
              				     					 & \multirow{2}{*}{nausea} &  type=ASPECTUAL \hspace{1cm} polarity=POS \\
                                                  					 &                                       &  degree=N/A \hspace{1cm} modality=HYPOTHETICAL\\
\cline{2-3}
\hline
\end{tabular}
\end{center}
\caption{An example of information extraction from clinical note. }
\label{tab:extract}
\end{table*}

To solve this task, the major challenge is how to precisely identify the spans (character offsets) of the event expressions from raw clinical notes. Traditional machine learning approaches usually build a supervised classifier with features generated by the Apache clinical Text Analysis and Knowledge Extraction System (cTAKES)~\footnote{Apache cTAKES is a natural language processing system for extraction of information from electronic medical record clinical free-text}. For example, BluLab system~\cite{velupillai2015blulab} extracted morphological(lemma), lexical(token), and syntactic(part-of-speech) features encoded from cTAKES. Although using the domain specific information extraction tools can improve the performance, learning how to use it well for clinical domain feature engineering is still very time-consuming.  In short, a simple and effective method that only leverage basic NLP modules and achieves high extraction performance is desired to save costs.

To address this challenge, we propose a deep neural networks based method, especially convolution neural network~\cite{Collobert2011}, to learn hidden feature representations directly from raw clinical notes. More specifically, one method first extract a window of surrounding words for the candidate word. Then, we attach each word with their part-of-speech tag and shape information as extra features. Then our system deploys a temporal convolution neural network to learn hidden feature representations. Finally, our system uses Multilayer Perceptron (MLP) to predict event spans. Note that we use the same model to predict event attributes.

\section{Constructing High Quality Training Dataset}
The major advantage of our system is that we only leverage NLTK~\footnote{http://www.nltk.org} tokenization and a POS tagger to preprocess our training dataset. When implementing our neural network based clinical information extraction system, we found it is not easy to construct high quality training data due to the noisy format of clinical notes. Choosing the proper tokenizer is quite important for span identification.  After several experiments, we found "RegexpTokenizer" can match our needs. This tokenizer can generate spans for each token via sophisticated regular expression like below, 
\begin{lstlisting}[language=Python]
nltk.tokenize.RegexpTokenizer
("\w+|\$[\d\.]+|\S+")
\end{lstlisting}

We then use "PerceptronTagger" as our part-of-speech tagger due to its fast tagging speed.  Note that when extracting context words, please make sure you deploy the same tokenization module instead of just splitting strings by space. 

\section{Neural Network Classifier}
Event span identification is the task of extracting character offsets of the expression in raw clinical notes. This subtask is quite important due to the fact that the event span identification accuracy will affect the accuracy of attribute identification. We first run our neural network classifier to identify event spans. Then, given each span, our system tries to identify attribute values. 

\subsection{Temporal Convolutional Neural Network}
The way we use temporal convlution neural network for event span and attribute classification is similar with the approach proposed by~\cite{Collobert2011}. Generally speaking, we can consider a word as represented by $K$ discrete features $w \in D^1 \times \cdots \times D^K $, where $D^K$ is the dictionary for the $k^{th}$ feature. In our scenario, we just use three features such as token mention, pos tag and word shape. Note that word shape features are used to represent the abstract letter pattern of the word by mapping lower-case letters to ``x'', upper-case to ``X'', numbers to ``d'', and retaining punctuation. We associate to each feature a lookup table. Given a word, a feature vector is then obtained by concatenating all lookup table outputs. Then a clinical snippet is transformed into a word embedding matrix. The matrix can be fed to further 1-dimension convolutional neural network and max pooling layers. Below we will briefly introduce core concepts of Convoluational Neural Network (CNN).

\subsubsection*{Temporal Convolution}
Temporal Convolution applies one-dimensional convolution over the input sequence. The one-dimensional convolution is an operation between a vector of weights $\bold{m} \in  \mathbb{R}^{m}$ and a vector of inputs viewed as a sequence $\bold{x} \in  \mathbb{R}^{n}$. The vector $\bold{m}$ is the \emph{filter} of the convolution. Concretely, we think of $\bold{x}$ as the input sentence and $\bold{x}_i \in \mathbb{R}$ as a single feature value associated with the $i$-th word in the sentence. The idea behind the one-dimensional convolution is to take the dot product of the vector $\bold{m}$ with each $m$-gram in the sentence $\bold{x}$ to obtain another sequence $\bold{c}$:
\begin{equation}
\bold{c}_j = \bold{m}^{T}\bold{x}_{j-m+1:j}\,.
\end{equation}

Usually, $\bold{x}_i$ is not a single value, but a $d$-dimensional word vector so that $\bold{x} \in \mathbb{R}^{d \times n}$. There exist two types of 1d convolution operations. One was introduced by~\cite{waibel1989phoneme} and also known as Time Delay Neural Networks (TDNNs). The other one was introduced by~\cite{Collobert2011}. In TDNN, weights $\bold{m} \in \mathbb{R}^{d \times m}$ form a matrix. Each row of $\bold{m}$ is convolved with the corresponding row of $\bold{x}$. In~\cite{Collobert2011} architecture, a sequence of length $n$ is represented as:
\begin{equation}
\bold{x}_{1:n} = \bold{x}_1 \oplus \bold{x}_2 \cdots \oplus \bold{x}_n\,,
\end{equation}
where $\oplus$ is the concatenation operation. In general, let $\bold{x}_{i:i+j}$ refer to the concatenation of words $\bold{x}_i, \bold{x}_{i+1}, \dots, \bold{x}_{i+j}$. A convolution operation involves a filter $\bold{w} \in \mathbb{R}^{hk}$, which is applied to a window of $h$ words to produce the new feature. For example, a feature $c_i$ is generated from a window of words $\bold{x}_{i:i+h-1}$ by:
\begin{equation}
c_i = f( \bold{w} \cdot \bold{x}_{i:i+h-1} + b )\,,
\end{equation}
where $b \in \mathbb{R}$ is a bias term and $f$ is a non-linear function such as the hyperbolic tangent. This filter is applied to each possible window of words in the sequence $\{ \bold{x}_{1:h},  \bold{x}_{2:h+1}, \dots, \bold{x}_{n-h+1:n} \}$ to produce the feature map:
\begin{equation}
\bold{c} = [ c_1, c_2, \dots, c_{n-h+1}]\,,
\end{equation}
where $\bold{c} \in \mathbb{R}^{n-h+1}$.

We also employ dropout on the penultimate layer with a constraint on $\ell_2$-norms of the weight vector. Dropout prevents co-adaptation of hidden units by randomly dropping out a proportion $p$ of the hidden units during forward-backpropagation. That is, given the penultimate layer $\bold{z} = [\hat{c_1}, \dots, \hat{c_m}]$, instead of using:
\begin{equation}
y = \bold{w} \cdot \bold{z} +b
\end{equation}
for output unit $y$ in forward propagation, dropout uses:
\begin{equation}
y = \bold{w} \cdot (\bold{z} \circ \bold{r}) +b \,,
\end{equation}
where $\circ$ is the element-wise multiplication operator and $\bold{r} \in \mathbb{R}^m$ is a masking vector of Bernoulli random variables with probability $p$ of being 1. Gradients are backpropagated only through the unmasked units. At test step, the learned weight vectors are scaled by $p$ such that $\hat{\bold{w}} = p\bold{w}$, and $\hat{\bold{w}}$ is used to score unseen sentences. We additionally constrain $l_2$-norms of the weight vectors by re-scaling $\bold{w}$ to have $|| \bold{w} ||_2 = s$ whenever $|| \bold{w} ||_2 > s$ after a gradient descent step.

\section{Experimental Results}

\subsection{Dataset}
We use the Clinical TempEval corpus~\footnote{http://alt.qcri.org/semeval2016/task12/index.php?id=data} as the evaluation dataset. This corpus was based on a set of 600 clinical notes and pathology reports from cancer patients at the Mayo Clinic. These notes were manually de-identified by the Mayo Clinic to replace names, locations, etc. with generic placeholders, but time expression were not altered. The notes were then manually annotated with times, events and temporal relations in clinical notes. These annotations include time expression types, event attributes and an increased focus on temporal relations. The event, time and temporal relation annotations were distributed separately from the text using the Anafora standoff format. Table~\ref{tab:data} shows the number of documents, event expressions in the training, development and testing portions of the 2016 THYME data. 

\begin{table}[tb]
\begin{center}
\large
\begin{tabular}{c|c|c|c}
\hline
Category & Train & Dev & Test\\
\hline
Documents & 293 & 147 & 151 \\
Events & 38872 & 20973 & 18989  \\
\hline
\end{tabular}
\end{center}
\caption{Number of documents, event expressions in the training, development and testing portions of the THYME data}
\label{tab:data}
\end{table}

\subsection{Evaluation Metrics}
All of the tasks were evaluated using the standard metrics of precision(P), recall(R) and $F_1$:

\begin{equation}
\begin{split}
P = \frac{| S \cap H |}{|S|} \\
R = \frac{| S \cap H |}{|H|} \\
F_1 =  \frac{2 \cdot P \cdot R}{ P + R} \\
\end{split}
\end{equation}
where $S$ is the set of items predicted by the system and $H$ is the set of items manually annotated by the humans. Applying these metrics of the tasks only requires a definition of what is considered an "item" for each task. For evaluating the spans of event expressions, items were tuples of character offsets. Thus, system only received credit for identifying events with exactly the same character offsets as the manually annotated ones. For evaluating the attributes of event expression types, items were tuples of (begin, end, value) where begin and end are character offsets and value is the value that was given to the relevant attribute. Thus, systems only received credit for an event attribute if they both found an event with correct character offsets and then assigned the correct value for that attribute~\cite{bethard2015semeval}. 

\subsection{Hyperparameters and Training Details}

\subsubsection*{Objective Function}
We want to maximize the likelihood of the correct class. This is equivalent to minimizing the negative log-likelihood (NLL). More specifically, the label $\hat{y}$ given the inputs $x_h$ is predicted by a softmax classifier that takes the hidden state $h_j$ as input:
\begin{equation}
\begin{split}
\hat{p}_\theta (y | x_h) &= softmax(W \cdot x_h + b) \\
\hat{y} &= \argmax_y \hat{p}_\theta (y | x_h)
\end{split}
\end{equation}
After that, the objective function is the negative log-likelihood of the true class labels $y^k$:
\begin{equation}
J(\theta) = -\frac{1}{m} \sum_{k=1}^m \log\hat{p}_\theta (y^k | x_h^k) + \frac{\lambda}{2} ||\theta ||_2^2\,,
\label{entcost}
\end{equation}
where $m$ is the number of training examples and the superscript $k$ indicates the $k$th example. 

\subsubsection*{Hyperparameters}
We use Lasagne~\footnote{\url{https://github.com/Lasagne/Lasagne}} deep learning framework. We first initialize our word representations using publicly available 300-dimensional Glove word vectors~\footnote{\url{http://nlp.stanford.edu/projects/glove/}}. We deploy CNN model with kernel width of 2, a filter size of 300, sequence length is $2*windows\_size + 1$, number filters is $seqlen-kw+1$, stride is 1, pool size is $seqlen-filter\_size+1$, cnn activation function is tangent, MLP activation function is sigmoid. MLP hidden dimension is 50. We initialize CNN weights using a uniform distribution. Finally, by stacking a softmax function on top, we can get normalized log-probabilities. Training is done through stochastic gradient descent over shuffled mini-batches with the AdaGrad update rule~\cite{Duchi2011}. The learning rate is set to 0.05. The mini-batch size is 100. The model parameters were regularized with a per-minibatch L2 regularization strength of $10^{-4}$. 

\subsection{Results and Discussions}
Table~\ref{tab:prf} shows results on the event expression tasks. Our initial submits RUN 4 and 5 outperformed the memorization baseline on every metric on every task. The precision of event span identification is close to the max report. However, our system got lower recall. One of the main reason is that our training objective function is accuracy-oriented.  Table~\ref{tab:prf2} shows results on the phase 2 subtask.

\begin{table*}[!htb]
\tiny
\begin{center}
\begin{tabular}{c|ccc|ccc|ccc|ccc|ccc}
\hline
\multicolumn{1}{c|}{} & \multicolumn{3}{c|}{span} & \multicolumn{3}{c|}{modality} & \multicolumn{3}{c|}{degree} & \multicolumn{3}{c|}{polarity} & \multicolumn{3}{c}{type}\\
\multicolumn{1}{c|}{Methods} & P & R & F1 & P & R & F1  & P & R & F1 & P & R & F1 & P & R & F1\\
\hline
Memorize & 0.878 & 0.834  & 0.855 & 0.810 & 0.770 & 0.789 & 0.874 & 0.831 & 0.852 & 0.812 & 0.772 & 0.792 & 0.855 & 0.813 & 0.833 \\
Ours RUN4 & 0.908 & 0.842  & 0.874 & 0.842 & 0.780 & 0.810 & 0.904 & 0.838 & 0.869 & 0.876 & 0.812 & 0.842 & 0.877 & 0.813 & 0.844 \\
Ours RUN5 & 0.900 & 0.850  & 0.874 & 0.837 & 0.790 & 0.813 & 0.896 & 0.845 & 0.870 & 0.861 & 0.813 & 0.836 & 0.869 & 0.820 & 0.844 \\
Median report & 0.887 & 0.846  & 0.874 & 0.830 & 0.780 & 0.810 & 0.882 & 0.838 & 0.869 & 0.868 & 0.813 & 0.839 & 0.854 & 0.813 & 0.844 \\
Max report & 0.915 & 0.891  & 0.903 & 0.866 & 0.843 & 0.855 & 0.911 & 0.887 & 0.899 & 0.900 & 0.875 & 0.887 & 0.894 & 0.870 & 0.882 \\
\hline
\end{tabular}
\end{center}
\caption{System performance comparison. Note that Run4 means the window size is 4, Run5 means the window size is 5}
\label{tab:prf}
\end{table*}

\begin{table}[!htb]
\small
\begin{center}
\begin{tabular}{|c|ccc|}
\hline
\multicolumn{1}{|c|}{} & \multicolumn{3}{c|}{DocTimeRel}\\
\multicolumn{1}{|c|}{Methods} & P & R & F1 \\
\hline
Memorize & \_ & 0.675  & \_  \\
Ours RUN5 & 0.788 & 0.788  & 0.788  \\
Ours RUN6 & 0.786 & 0.786  & 0.786  \\
Median report & \_ & 0.724  & \_  \\
Max report & \_ & 0.843  & \_  \\
\hline
\end{tabular}
\end{center}
\caption{Phase 2: DocTimeRel}
\label{tab:prf2}
\end{table}

\section{Conclusions}
In this paper, we introduced a new clinical information extraction system that only leverage deep neural networks to identify event spans and their attributes from raw clinical notes. We trained deep neural networks based classifiers to extract clinical event spans. Our method attached each word to their part-of-speech tag and shape information as extra features. We then hire temporal convolution neural network to learn hidden feature representations.  The entire experimental results demonstrate that our approach consistently outperforms the existing baseline methods on standard evaluation datasets.

Our research proved that we can get competitive results without the help of a domain specific feature extraction toolkit, such as cTAKES. Also we only leverage basic natural language processing modules such as tokenization and part-of-speech tagging. With the help of deep representation learning, we can dramatically reduce the cost of clinical information extraction system development. 

\bibliographystyle{naaclhlt2016}
\bibliography{Bibliographies}

\end{document}